\documentclass{article}

\PassOptionsToPackage{numbers,compress}{natbib}
\usepackage[preprint]{neurips_2026}

\usepackage[utf8]{inputenc}
\usepackage[T1]{fontenc}
\usepackage{amsmath}
\usepackage{amsfonts}
\usepackage{graphicx}
\usepackage{booktabs}
\usepackage{tabularx}
\usepackage{pifont}
\usepackage{microtype}
\usepackage{xcolor}
\usepackage{url}
\usepackage{xurl}
\usepackage{hyperref}
\hypersetup{hidelinks}
\usepackage{makecell}
\usepackage{multirow}
\usepackage{float}

\title{TRACE: A Reproducible Benchmark for Electricity Price Forecasting
with Official Operational Text\thanks{Accepted to the NeurIPS 2026 Workshop on Foundation Models for Time Series (FMTS).}}

\workshoptitle{Foundation Models for Temporal Systems: From Forecasting to World Modeling}

\author{%
  Xinyi Yi \\
  Department of Engineering\\
  University of Cambridge\\
  \texttt{xy343@cam.ac.uk} \\
  \And
  Zhangdie Yuan \\
  Department of Computer Science\\
  University of Cambridge\\
  \texttt{zy317@cam.ac.uk} \\
  \And
  Ioannis Lestas \\
  University of Cambridge\\
  \texttt{icl20@cam.ac.uk} \\
}

\begin{document}

\maketitle
\begin{abstract}
Electricity price forecasting (EPF) supports scheduling, bidding, and risk management in electricity markets, yet existing benchmarks focus mainly on numerical inputs, leaving the forecasting value of forecast-time textual context insufficiently evaluated. We introduce TRACE, a reproducible benchmark of 7,300 zone--day instances
pairing prices from five zones in a major U.S. market with official operational
text available at the forecast cutoff. TRACE reconstructs official operational
text at each cutoff, preventing post-cutoff information leakage. We evaluate TRACE for semantic alignment and forecasting value. Semantic
assessments align with central movement and both tail risks in ground-truth prices,
most consistently for upper-tail price risk. Forecasting value is reflected in a median 7.4\% reduction in upper-tail
pinball loss across time-series foundation models. A controlled cross-day text-mismatch ablation
reverses the gains, falling below the no-text baseline.
\end{abstract}


\section{Introduction}

Electricity price forecasting (EPF) supports bidding, scheduling, and risk
management in electricity markets~\citep{joskow2007reliability,hogan2013electricity}.
Most methods rely on historical prices and other numerical inputs,
possibly augmented with demand, weather, fuel, or calendar
variables~\citep{3737916.3737931}. While these inputs provide strong baselines, the forecasting value of forecast-time textual context remains underexplored.

System operators publish official operational text on scarcity,
constraints, weather stress, and operational actions. These official
operational records can signal changing conditions before they appear in
recent prices. For example, on 6~January 2022, PJM, a major U.S. electricity
market operator~\citep{pjm_introdu}, \footnote{PJM coordinates the wholesale electricity market and bulk power system across 13 states and the District of Columbia.} issued a Cold Weather Alert before 7~January day-ahead prices were determined;
in COMED,\footnote{COMED (Commonwealth Edison) is a PJM market zone covering
northern Illinois, including the Chicago area.} nine hourly prices
subsequently exceeded the prior 14-day maximum. This case illustrates the
potential for official operational text to provide forecasting value beyond
recent price history. Textual context has likewise shown forecasting value
beyond numerical inputs in equities~\citep{tetlock2007giving} and
commodities~\citep{smales2014news}. Electricity markets provide a natural
setting for studying textual context beyond numerical inputs: prices exhibit recurring calendar patterns~\citep{weron2014electricity,lago2021forecasting}
yet respond sharply to operational stress.

Recent multimodal time-series benchmarks pair numerical inputs with both
naturally occurring and constructed textual context, but do not target EPF.
Time-MMD, for example, spans agriculture, climate, gasoline prices, health,
and traffic~\citep{liu2024time}, whereas other benchmarks construct
forecasting tasks and textual context so that the context is informative by
design~\citep{williams2024context}. Related work also studies irregular
multimodal observations in healthcare, climate, and
finance~\citep{chang2026time}, and data-leakage-aware context
construction~\citep{xu2025fidel}. In EPF, concurrent work
NSW-EPNews~\citep{bi2025nswepnews} reports limited gains from textual context,
but uses third-party market commentary and GPT-4o-generated summaries,
complicating forecast-time reconstruction and making benchmark construction
harder to reproduce.

\begin{figure}[htbp!]
\centering
\includegraphics[width=\linewidth]{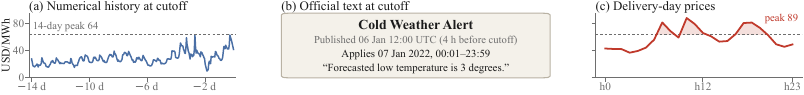}
\caption{
TRACE instance from COMED (target: 7~January 2022).
A Cold Weather Alert was issued four hours before the forecast cutoff (b);
9 prices exceeded the prior 14-day maximum (a,c).
}
\label{fig:examplerift}
\end{figure}

We introduce TRACE (Temporal Reconstruction of Available Context for
Electricity Price Forecasting), to our knowledge the first reproducible EPF
benchmark pairing prices~\citep{pjm_dataminer} with official operational text ~\citep{pjm_emergency_procedures} reconstructed at each
forecast cutoff. Across 1,460 days and five
zones, TRACE uses publication time, cutoff-time cancellation status,
and applicability intervals announced at publication to construct 7,300
zone--day instances without post-cutoff information leakage. Its construction
applies unchanged as new official operational records arrive.
Figure~\ref{fig:examplerift} shows one instance.

TRACE makes three contributions. First, it provides a reproducible benchmark
pairing prices from five PJM zones with official operational text reconstructed
at the forecast cutoff. Second, it evaluates semantic alignment: semantic assessments align with
central movement and both tail risks, most consistently
for upper-tail price risk. Third, it evaluates forecasting value: forecast-time
textual context reduces upper-tail pinball loss by a median $7.4\%$ across
time-series foundation models, while a controlled cross-day text-mismatch
ablation reverses gains, falling below the no-text baseline.

\section{TRACE benchmark construction}
\label{sec:benchmark}

\begin{figure}[t!]
    \centering
    \includegraphics[width=\linewidth]{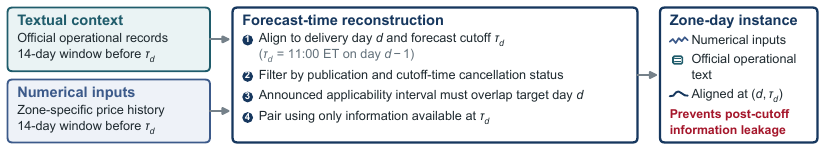}
   \caption{
TRACE construction.}
    \label{fig:rift_pipeline}
\end{figure}

For delivery day $d$, the forecast cutoff $\tau_d$ is 11:00~ET on day $d-1$,
when PJM's day-ahead bid period closes~\citep{pjm_manual11}. TRACE
deterministically aligns zone-specific price history and official operational
text available at $\tau_d$ with the target for day $d$.
Figure~\ref{fig:rift_pipeline} summarizes the construction.

\textbf{Deterministic forecast-time textual context reconstruction.}
TRACE uses PJM's Emergency Procedures archive~\citep{pjm_emergency_procedures}
for precise forecast-time reconstruction and broad coverage
(Appendix~\ref{app:source-selection}). We fix a 14-day window before evaluation,
retaining $87.6\%$ of unlimited-lookback coverage while limiting stale context
(Appendix~\ref{app:window-reconstruction}). Records are filtered by publication
and cutoff-time cancellation status; announced applicability intervals must
overlap the target day. Thus, only information available at the forecast cutoff
enters the benchmark, preventing post-cutoff information leakage.

\textbf{Numerical inputs.}
Numerical inputs use the same 14-day window of zone-specific
prices~\citep{pjm_dataminer}. Following common DST normalization
~\citep{hauzenberger2025sparse}, spring-forward gaps are interpolated and one
fall-back observation retained, yielding 336 hourly inputs.

\textbf{Cross-modal alignment and benchmark splits.}
Cross-modal inputs align by $(d,\tau_d)$: prices use the preceding 14-day prices,
while text includes records published and uncancelled by $\tau_d$ and
applicable to $d$. TRACE contains 7,300 five-zone instances over 1,460 days
(2021-10-02--2025-09-30), chronological split 821/366/273 train/validation/test; 702 records cover
400/146/131 days.

\section{Evaluating semantic alignment and forecasting value in TRACE}
\label{sec:evaluation}

\subsection{Semantic alignment of textual context}
\label{sec:semantic-alignment}

We test whether official operational text available at the forecast cutoff
contains information about subsequent prices. For each delivery day $d$, an
LLM semantic assessor reads all retained official operational text available by its forecast
cutoff $\tau_d$ and outputs three assessments: central movement $\hat C_d$, upper-tail
price risk $\hat S_d^U$, and lower-tail price risk $\hat S_d^L$. All use
$\{-1,0,+1\}$, denoting \textsc{down}/\textsc{neutral}/\textsc{up} for
$\hat C_d$ and \textsc{decrease}/\textsc{no clear change}/\textsc{increase}
for the tails.

\textbf{Ground-truth central movement and tail risks.}
Let $\mathcal D$ be training and validation days with textual context. Let
$y_{d,h}$ be the ground-truth price at hour $h$ on day $d$, and
$\bar y_d$ its daily mean. Using the previous-day reference, define
$D_d=\bar y_d-\bar y_{d-1}$, $r_{d,h}=y_{d,h}-y_{d-1,h}$, and
$\delta=0.1\,\operatorname{med}_{d\in\mathcal D}|D_d|$.
The ground-truth central-movement direction and day-level tail-risk levels are
\[
C_d=
\begin{cases}
+1, & D_d>\delta,\\
0, & |D_d|\le\delta,\\
-1, & D_d<-\delta,
\end{cases}
\qquad
U_d=Q_{0.9}(\{r_{d,h}\}_h),\qquad
L_d=-Q_{0.1}(\{r_{d,h}\}_h).
\]
To account for recurring calendar patterns~\citep{weron2014electricity,lago2021forecasting}, we repeat the construction using
four-week same-weekday medians,
$\operatorname{med}_{j=1}^{4}\bar y_{d-7j}$ and
$\operatorname{med}_{j=1}^{4}y_{d-7j,h}$, in place of
$\bar y_{d-1}$ and $y_{d-1,h}$, respectively.

\textbf{Alignment metrics.}
We compare the LLM assessments with ground-truth central-movement direction $C_d$
and upper- and lower-tail risks $U_d$ and $L_d$.
Central alignment uses balanced accuracy $A_C$ (chance $1/3$), while tail
alignment uses Spearman correlation~\citep{spearman1961proof} across days, $A_U$ and $A_L$ (null $0$):
\[
\begin{aligned}
A_C=&\frac{1}{3}\sum_{c\in\{-1,0,+1\}}
\Pr_{d\in\mathcal D}(\hat C_d=c\mid C_d=c),\\
A_U=&\rho_S\!\left((\hat S_d^U)_{d\in\mathcal D},(U_d)_{d\in\mathcal D}\right),
A_L=\rho_S\!\left((\hat S_d^L)_{d\in\mathcal D},(L_d)_{d\in\mathcal D}\right).
\end{aligned}
\]
Here, $A_C\in[0,1]$ and $A_U,A_L\in[-1,1]$, with larger values indicating
stronger positive alignment. 


\textbf{Experimental results.}
We use Claude Opus~5~\citep{anthropic2026opus5} and DeepSeek
V4-Flash~\citep{deepseekai2026deepseekv4} as LLM semantic
assessors with the same frozen prompt (Appendix~\ref{app:pjm-semantic}).
Assessments are frozen and shared across zones; alignment is zone-specific.
Significance uses $1{,}000$ joint day-level permutations, one-sided $p$-values,
and Holm--Bonferroni correction over three dimensions, two references within
each zone--LLM pair~\citep{holm1979simple}. Positive, significant alignment
($A_C>1/3$ or $A_U,A_L>0$; adjusted $p<0.05$) occurs in 17/20 central,
20/20 upper-tail, and 16/20 lower-tail settings. Upper-tail risk is most consistent, matching the strongest and most consistent
forecasting value of textual context below (Table~\ref{tab:audit}).

\begin{table}[H]
\centering
\vspace{-2pt}
\caption{Cross-zone semantic alignment.
Full results are in Appendix~\ref{app:alignment-full}.}
\label{tab:audit}

\scriptsize
\renewcommand{\arraystretch}{0.78}

\begin{minipage}[t]{0.50\linewidth}
\centering
\textit{A. Positive, Holm-significant alignment}\\[-1pt]
{\scriptsize Positive: $A_C>1/3$, $A_U,A_L>0$; significant: Holm $p<0.05$}\\
\begin{tabular*}{\linewidth}{@{\extracolsep{\fill}}lccc@{}}
\toprule
Zone & Central $A_C$ & Upper $A_U$ & Lower $A_L$ \\
\midrule
COMED & 2/4 & \textbf{4/4} & 2/4 \\
AEP   & 3/4 & \textbf{4/4} & 2/4 \\
DOM   & 4/4 & \textbf{4/4} & 4/4 \\
PSEG  & 4/4 & \textbf{4/4} & 4/4 \\
BGE   & 4/4 & \textbf{4/4} & 4/4 \\
\midrule
All   & 17/20 & \textbf{20/20} & 16/20 \\
\bottomrule
\end{tabular*}
\end{minipage}\hfill
%
\begin{minipage}[t]{0.48\linewidth}
\centering
\textit{B. Upper-tail alignment $A_U$}\\[-1pt]
{\scriptsize All entries: $A_U>0$, Holm $p<0.05$}\\
\begin{tabular*}{\linewidth}{@{\extracolsep{\fill}}lcccc@{}}
\toprule
& \multicolumn{2}{c}{Claude} & \multicolumn{2}{c}{DeepSeek} \\
\cmidrule(lr){2-3}\cmidrule(lr){4-5}
Zone & Prev. & Same-wd. & Prev. & Same-wd. \\
\midrule
COMED & 0.164 & 0.429 & 0.177 & 0.415 \\
AEP   & 0.247 & 0.402 & 0.263 & 0.390 \\
DOM   & 0.282 & 0.363 & 0.301 & 0.371 \\
PSEG  & 0.347 & 0.449 & 0.374 & 0.432 \\
BGE   & 0.293 & 0.398 & 0.284 & 0.398 \\
\bottomrule
\end{tabular*}
\end{minipage}

\end{table}

\subsection{Forecasting value of textual context}
\label{sec:forecasting}

We evaluate TRACE forecasting in two settings: \textbf{direct text fusion}
~\citep{liu2024time,huang2025crossmoe} combines encoded official operational
text with numerical features, while \textbf{structured semantic fusion}
~\citep{li2026text} uses semantic assessments. For
$\tau\in\{0.1,0.5,0.9\}$, we use pinball loss
~\citep{lim2021temporal,cornell2024probabilistic},
$\mathrm{PB}_\tau=|\Omega|^{-1}\sum_{(d,h)\in\Omega}
(\tau-\mathbf{1}\{y_{d,h}<q_{d,h}^{\tau}\})(y_{d,h}-q_{d,h}^{\tau})$,
where $\Omega$ indexes test day--hour pairs, $y_{d,h}$ is the ground-truth
price, and $q_{d,h}^{\tau}$ the predicted $\tau$-quantile; lower is better.
Seven zero-shot time-series foundation models serve as baselines:
TimesFM~2.5~\citep{das2024decoder}, TimesFM~3.0~\citep{jain2026timesfm3},
Chronos-Bolt-Tiny/Base~\citep{ansari2024chronos},
Chronos-2~\citep{ansari2025chronos2}, Moirai-2~\citep{liu2025moirai2},
and Sundial~\citep{liu2025sundial}, using 336 hours of history to forecast
24 hours.

\begin{figure}[htbp]
\centering
\includegraphics[width=\linewidth]{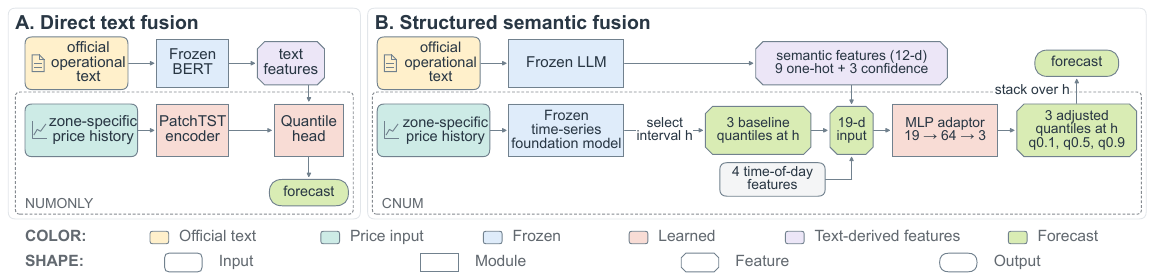}
\vspace{-0.3cm}
\caption{
Direct text fusion (A) and structured semantic fusion (B). 
}
\label{fig:fusion}
\end{figure}

For \textbf{direct text fusion} (Figure~\ref{fig:fusion}A), following
MM-TSFlib~\citep{liu2024time}, trainable PatchTST~\citep{nie2022time} encodes the 14-day
zone-specific price history and frozen BERT~\citep{devlin2019bert} encodes
chronologically concatenated official operational text. The resulting
representations feed a quantile head trained jointly (except BERT) with
equal-weight pinball loss over $\tau\in\{0.1,0.5,0.9\}$; results average five
seeds. \textsc{NumOnly} is trained identically with text features zeroed; gains
are reported relative to \textsc{NumOnly}.

For \textbf{structured semantic fusion} (Figure~\ref{fig:fusion}B),
frozen LLMs provide a 12-dimensional vector of three one-hot semantic assessments
and confidence scores. For each forecast hour, the MLP takes this vector, three
baseline quantile forecasts from the frozen time-series foundation model, and
four time-of-day features as 19 inputs, and outputs quantile forecasts. The $19\!\rightarrow\!64\!\rightarrow\!3$ adaptor is
trained with equal-weight pinball loss while preserving quantile order.
\textsc{Cnum} is trained identically with the vector zeroed; gains over
\textsc{Cnum} isolate the semantic contribution beyond numerical recalibration.

Table~\ref{tab:forecasting-main} summarizes upper-tail forecasting gains.
For each model family and zone, it reports the member with the lowest frozen
$\mathrm{PB}_{0.9}$; full seven-model results are in
Appendix~\ref{app:forecasting-full}. Direct text fusion reduces
$\mathrm{PB}_{0.9}$ in all five zones by \textbf{3.7--9.1\%} versus
\textsc{NumOnly}. At $\tau=0.9$, structured semantic fusion improves all
\textbf{70/70} zone--model--LLM settings over \textsc{Cnum}, with a median
\textbf{7.4\%} reduction (range $1.4$--$19.3\%$), and by
\textbf{11.3\%} versus the frozen baselines (range $2.8$--$24.3\%$);
at $\tau=0.5$, \textbf{64/70} settings improve. Correct pairing beats the
mean over 1,000 context re-pairings in all \textbf{5/5} direct-text zones and
\textbf{70/70} semantic settings, with median gains of \textbf{20.7\%} and
\textbf{13.9\%}, respectively.

\vspace{-0.3cm}

\begin{table}[htbp]
\centering

\caption{
Forecasting results of direct text fusion and structured semantic fusion.}
\label{tab:forecasting-main}

\scriptsize
\renewcommand{\arraystretch}{0.78}

\begin{minipage}[t]{0.66\linewidth}
\vspace{-2pt}
\centering

\textit{A. Direct text fusion at $\tau=0.9$}\\[-1pt]

\begin{tabular*}{\linewidth}{
@{\extracolsep{\fill}}
lrrr
@{}
}
\toprule
Zone
& Direct
& $\downarrow\%$ vs.\ \textsc{NumOnly}
& $\downarrow\%$ vs.\ Shuffle
\\
\midrule
COMED & \textbf{4.8973} & \textbf{6.4\%} & \textbf{19.6\%} \\
AEP   & \textbf{4.4380} & \textbf{7.2\%} & \textbf{20.6\%} \\
DOM   & \textbf{8.4872} & \textbf{7.9\%} & \textbf{20.7\%} \\
PSEG  & \textbf{5.2073} & \textbf{3.7\%} & \textbf{23.8\%} \\
BGE   & \textbf{6.9271} & \textbf{9.1\%} & \textbf{27.4\%} \\
\bottomrule
\end{tabular*}

\end{minipage}
\hfill
\begin{minipage}[t]{0.31\linewidth}
\vspace{0pt}
\centering

\textit{B. Semantic fusion}\\[-1pt]

\begin{tabular*}{\linewidth}{
@{\extracolsep{\fill}}
cr
@{}
}
\toprule
$\tau$ & Improved settings \\
\midrule
0.9 & \textbf{70/70 (100\%)} \\
0.5 & 64/70 (91\%) \\
0.1 & 24/70 (34\%) \\  
\bottomrule
\end{tabular*}

\end{minipage}

\vspace{0pt}

\renewcommand{\arraystretch}{0.58}

\textit{C. Structured semantic fusion at $\tau=0.9$}\\[-1pt]

\begin{tabular*}{\linewidth}{
@{\extracolsep{\fill}}
lp{59pt}
cc
cc
cc
@{}
}
\toprule
&
&
&
&
\multicolumn{2}{c}{Claude Opus 5}
&
\multicolumn{2}{c}{DeepSeek V4-Flash}
\\
\cmidrule(lr){5-6}
\cmidrule(lr){7-8}

Zone
& Baseline
& Frozen
& \textsc{Cnum}
& \textbf{Fusion} ($\downarrow\%$ vs.\ \textsc{Cnum})
& Shuffle
& \textbf{Fusion} ($\downarrow\%$ vs.\ \textsc{Cnum})
& Shuffle
\\
\midrule


COMED
& \textbf{TimesFM 3.0}
& 3.722
& 3.605
& \textbf{3.403 (5.6\%)}
& 3.861
& \textbf{3.416 (5.2\%)}
& 3.872
\\

&
\textbf{Chronos-Bolt-Base}
& 3.667
& 3.530
& \textbf{3.396 (3.8\%)}
& 3.713
& \textbf{3.410 (3.4\%)}
& 3.761
\\

&
\textbf{Moirai-2}
& 3.863
& 3.765
& \textbf{3.465 (8.0\%)}
& 4.092
& \textbf{3.506 (6.9\%)}
& 4.163
\\

&
\textbf{Sundial}
& 4.766
& 3.862
& \textbf{3.606 (6.6\%)}
& 4.069
& \textbf{3.626 (6.1\%)}
& 3.975
\\


AEP
& \textbf{TimesFM 3.0}
& 3.871
& 3.770
& \textbf{3.432 (9.0\%)}
& 3.803
& \textbf{3.508 (6.9\%)}
& 3.805
\\

&
\textbf{Chronos-Bolt-Base}
& 3.791
& 3.628
& \textbf{3.330 (8.2\%)}
& 3.778
& \textbf{3.376 (7.0\%)}
& 3.825
\\

&
\textbf{Moirai-2}
& 3.941
& 3.822
& \textbf{3.533 (7.5\%)}
& 3.947
& \textbf{3.535 (7.5\%)}
& 3.966
\\

&
\textbf{Sundial}
& 4.643
& 4.057
& \textbf{3.823 (5.8\%)}
& 4.139
& \textbf{3.908 (3.7\%)}
& 4.121
\\


DOM
& \textbf{TimesFM 3.0}
& 6.956
& 6.914
& \textbf{6.573 (4.9\%)}
& 7.619
& \textbf{6.758 (2.3\%)}
& 7.938
\\

&
\textbf{Chronos-2}
& 7.225
& 7.026
& \textbf{6.521 (7.2\%)}
& 7.243
& \textbf{6.604 (6.0\%)}
& 7.403
\\

&
\textbf{Moirai-2}
& 7.316
& 6.923
& \textbf{6.490 (6.3\%)}
& 7.549
& \textbf{6.617 (4.4\%)}
& 7.910
\\

&
\textbf{Sundial}
& 8.719
& 7.544
& \textbf{7.180 (4.8\%)}
& 8.130
& \textbf{7.334 (2.8\%)}
& 8.395
\\


PSEG
& \textbf{TimesFM 3.0}
& 4.158
& 4.099
& \textbf{3.446 (15.9\%)}
& 4.459
& \textbf{3.517 (14.2\%)}
& 4.244
\\

&
\textbf{Chronos-2}
& 3.986
& 3.689
& \textbf{3.300 (10.5\%)}
& 4.265
& \textbf{3.427 (7.1\%)}
& 4.368
\\

&
\textbf{Moirai-2}
& 4.173
& 4.040
& \textbf{3.432 (15.0\%)}
& 4.357
& \textbf{3.492 (13.6\%)}
& 4.251
\\

&
\textbf{Sundial}
& 5.151
& 4.456
& \textbf{3.928 (11.8\%)}
& 4.851
& \textbf{3.972 (10.9\%)}
& 4.806
\\


BGE
& \textbf{TimesFM 3.0}
& 6.131
& 6.082
& \textbf{5.524 (9.2\%)}
& 6.706
& \textbf{5.670 (6.8\%)}
& 6.931
\\

&
\textbf{Chronos-2}
& 5.986
& 5.902
& \textbf{5.488 (7.0\%)}
& 6.365
& \textbf{5.534 (6.2\%)}
& 6.462
\\

&
\textbf{Moirai-2}
& 6.454
& 6.255
& \textbf{5.590 (10.6\%)}
& 7.025
& \textbf{5.744 (8.2\%)}
& 7.366
\\

&
\textbf{Sundial}
& 7.634
& 6.644
& \textbf{5.956 (10.4\%)}
& 7.136
& \textbf{6.009 (9.6\%)}
& 7.278
\\

\bottomrule
\end{tabular*}

\end{table}

\section{Discussion and outlook} \label{sec:conclusion} 
We propose TRACE, the first reproducible EPF benchmark with official operational text. Semantic assessments show strongest alignment with upper-tail risk, where forecasting gains are strongest.

\bibliographystyle{unsrtnat}
\bibliography{refs}

\clearpage
\appendix
\section{Benchmark construction details}

\subsection{Official operational text source selection}
\label{app:source-selection}

We select the text source before any semantic or forecasting evaluation.
For delivery day $d$, let $\tau_d$ denote the forecast cutoff and let $p_i$
be the publication time of record $i$. Under unlimited lookback, let
\(p_d^\star=\max_{i:p_i<\tau_d}p_i\)
denote the latest publication available before the forecast cutoff for delivery day \(d\),
and define \(a_d=\tau_d-p_d^\star\) as its age.
Thus, ``Within \(k\) days'' is the fraction of delivery days whose latest
pre-cutoff publication is no more than \(k\) days old.

Both Emergency Procedures~\cite{pjm_emergency_procedures} and Inside
Lines~\cite{pjm_inside_lines} have at least one pre-cutoff publication for all
1{,}460 delivery days. Emergency Procedures is more recent: its median
$a_d$ is 1.15 days, compared with 3.00 days for Inside Lines
(Table~\ref{tab:source-selection}). It also provides the publication and
cancellation timestamps required for forecast-time reconstruction. Inside
Lines provides update timestamps but does not preserve superseded text, so its
state at an earlier forecast cutoff cannot be reconstructed. We therefore use
the PJM Emergency Procedures archive as the source of official operational text.

\begin{table}[htbp]
\centering
\caption{
Recency of candidate PJM text sources under unlimited lookback.
}
\label{tab:source-selection}
\small
\begin{tabular}{lrrrr}
\toprule
Source & Median age & Within 1 day & Within 3 days & Within 7 days \\
\midrule
Emergency Procedures & 1.15 days & 44.5\% & 77.4\% & 94.9\% \\
Inside Lines          & 3.00 days & 21.3\% & 50.1\% & 79.9\% \\
\bottomrule
\end{tabular}
\end{table}

\subsection{Context window and coverage}
\label{app:window-reconstruction}

After selecting Emergency Procedures, we fix the textual context window before
any semantic or forecasting evaluation. Let $p_{\min}$ be the earliest
publication in the collected archive (2021-09-04). For delivery day $d$ with
forecast cutoff $\tau_d$, a $W$-day window considers records published in
\[
[\max\{p_{\min},\,\tau_d-W\},\,\tau_d).
\]
The full-archive setting removes the finite lookback limit and therefore uses
$[p_{\min},\tau_d)$. Thus, benchmark days near the start of the evaluation
period may use records published before the first delivery day.
Within each window, records are filtered by publication and cutoff-time
cancellation status; if an applicability interval was announced at publication,
it must overlap delivery day $d$. Let $\mathcal{Z}_d(W)$ denote the retained
records. A day is context-bearing when $|\mathcal{Z}_d(W)|>0$.

Table~\ref{tab:window-sensitivity} shows the resulting coverage. The full
archive yields 773 context-bearing days. A 14-day window retains 677 of them
($87.6\%$); extending to 28 days adds only 42 days while doubling the maximum
record age. We therefore set $W=14$ days.

\begin{table}[htbp]
\centering
\caption{
Sensitivity of forecast-time context coverage to the lookback window.
Coverage is relative to the 773 context-bearing days obtained from the full
archive.
}
\label{tab:window-sensitivity}
\small
\begin{tabular}{lrr}
\toprule
Context window & Context-bearing days & Coverage \\
\midrule
1 day        & 326 & 42.2\% \\
7 days       & 621 & 80.3\% \\
14 days      & 677 & 87.6\% \\
28 days      & 719 & 93.0\% \\
Full archive & 773 & 100.0\% \\
\bottomrule
\end{tabular}
\end{table}

\section{Semantic evaluation details}

\subsection{Frozen semantic assessment prompt}
\label{app:pjm-semantic}

Claude Opus~5~\citep{anthropic2026opus5} and DeepSeek
V4-Flash~\citep{deepseekai2026deepseekv4} independently assess all 546
context-bearing training and validation days using the same frozen prompt.
For each day, an LLM receives only the official operational text retained by
forecast-time reconstruction within the 14-day context window, including record
type, publication time, announced applicability start time, and body text. It
returns three categorical semantic assessments and corresponding confidence
scores: central movement, upper-tail price risk, and lower-tail price risk.
The LLM receives no price history, numerical forecasts, ground-truth prices, or
previous-day or same-weekday reference values. All outputs are frozen before
alignment with ground-truth prices is computed.

\paragraph{Frozen prompt.}
The prompt below preserves the semantic task and label definitions while
omitting model-specific API formatting and output boilerplate.

\begin{quote}
You are evaluating official electricity-market information for a day-ahead
forecast. Make three independent assessments of what the supplied context implies
for the target delivery day.

You have no access to any ground-truth price, historical price, numerical forecast,
model output, or baseline. Assess only the economic or operational pressure
conveyed by the supplied context. Do not use generic seasonal, weekday, or
calendar priors.

\textbf{Central price direction.}
Does the context imply net upward, net downward, or no clear directional pressure
on the price level for the target delivery day?

\textsc{UP}: net upward pressure.\\
\textsc{DOWN}: net downward pressure.\\
\textsc{NEUTRAL\_MIXED}: no sufficiently clear net direction, materially
offsetting effects, or evidence too weak or indirect to support a direction.

\textbf{Upper-tail risk change.}
Does the context make unusually high price outcomes on the target delivery day
more likely, less likely, or neither?

\textsc{INCREASE}, \textsc{DECREASE}, or \textsc{NO\_CLEAR\_CHANGE}.

\textbf{Lower-tail risk change.}
Does the context make unusually low price outcomes on the target delivery day
more likely, less likely, or neither?

\textsc{INCREASE}, \textsc{DECREASE}, or \textsc{NO\_CLEAR\_CHANGE}.

Treat the three assessments independently: do not derive either tail assessment
from the central direction, or vice versa. Consider the magnitude, timing,
persistence, freshness, and operational relevance of the supplied events. When
the evidence is weak or conflicting, use the no-clear-signal labels rather than
forcing a prediction.

Return one label and confidence score for each assessment.
\end{quote}

\subsection{Full five-zone semantic alignment}
\label{app:alignment-full}

Table~\ref{tab:alignment-full} reports semantic alignment for every zone--LLM
pair. For each pair, we jointly permute
$(\hat C_d,\hat S_d^U,\hat S_d^L)$ across days $1{,}000$ times while holding
ground-truth prices fixed, and recompute
$A_C^{\mathrm{prev}},A_U^{\mathrm{prev}},A_L^{\mathrm{prev}},
A_C^{\mathrm{cal}},A_U^{\mathrm{cal}},A_L^{\mathrm{cal}}$.
The resulting permutation $p$-values are Holm--Bonferroni corrected across
these six statistics within each zone--LLM pair; the minimum attainable
corrected $p$-value is $0.006$. Alignment uses 546 context-bearing training and
validation days for the previous-day reference and 542 for the same-weekday
reference, as four days lack sufficient prior same-weekday observations.
Because textual context is shared across zones, semantic assessments are fixed
across zones, while ground-truth quantities and alignment statistics are
zone-specific.

Upper-tail alignment is positive and significant for every zone--LLM pair under
both references. Central movement is significant under both references for
seven of ten pairs, and lower-tail price risk for six of ten. Under the
previous-day reference, central movement and lower-tail price risk are
nonsignificant for COMED, and lower-tail price risk is also nonsignificant for
AEP. Upper-tail price risk is the only semantic dimension significant under
both references for every zone--LLM pair.

\begin{table}[htbp]
\centering
\caption{Five-zone semantic alignment on context-bearing training and validation
days. Entries are observed statistics with Holm-corrected permutation
$p$-values in parentheses; bold denotes positive alignment with
$p_{\mathrm{Holm}}<0.05$. $A_C$ is balanced accuracy (chance $1/3$), while
$A_U$ and $A_L$ are Spearman correlations (null $0$).}
\label{tab:alignment-full}
\scriptsize
\setlength{\tabcolsep}{3pt}
\begin{tabular}{llcccccc}
\toprule
& & \multicolumn{2}{c}{Central $A_C$} & \multicolumn{2}{c}{Upper tail $A_U$}
& \multicolumn{2}{c}{Lower tail $A_L$} \\
\cmidrule(lr){3-4}\cmidrule(lr){5-6}\cmidrule(lr){7-8}
Zone & LLM & prev.\ day & same wkday & prev.\ day & same wkday & prev.\ day & same wkday \\
\midrule
COMED & Claude & 0.369 (0.140) & \textbf{0.402} (0.012) & \textbf{0.164} (0.006) & \textbf{0.429} (0.006) & $-0.008$ (0.661) & \textbf{0.339} (0.006) \\
 & DeepSeek & 0.361 (0.250) & \textbf{0.394} (0.012) & \textbf{0.177} (0.006) & \textbf{0.415} (0.006) & 0.014 (0.457) & \textbf{0.341} (0.006) \\
\addlinespace[1pt]
AEP & Claude & \textbf{0.413} (0.006) & 0.384 (0.060) & \textbf{0.247} (0.006) & \textbf{0.402} (0.006) & 0.033 (0.272) & \textbf{0.336} (0.006) \\
 & DeepSeek & \textbf{0.413} (0.006) & \textbf{0.399} (0.006) & \textbf{0.263} (0.006) & \textbf{0.390} (0.006) & 0.069 (0.069) & \textbf{0.310} (0.006) \\
\addlinespace[1pt]
DOM & Claude & \textbf{0.411} (0.006) & \textbf{0.419} (0.006) & \textbf{0.282} (0.006) & \textbf{0.363} (0.006) & \textbf{0.107} (0.012) & \textbf{0.330} (0.006) \\
 & DeepSeek & \textbf{0.412} (0.006) & \textbf{0.429} (0.006) & \textbf{0.301} (0.006) & \textbf{0.371} (0.006) & \textbf{0.139} (0.006) & \textbf{0.317} (0.006) \\
\addlinespace[1pt]
PSEG & Claude & \textbf{0.419} (0.006) & \textbf{0.426} (0.006) & \textbf{0.347} (0.006) & \textbf{0.449} (0.006) & \textbf{0.139} (0.006) & \textbf{0.316} (0.006) \\
 & DeepSeek & \textbf{0.420} (0.006) & \textbf{0.422} (0.006) & \textbf{0.374} (0.006) & \textbf{0.432} (0.006) & \textbf{0.174} (0.006) & \textbf{0.314} (0.006) \\
\addlinespace[1pt]
BGE & Claude & \textbf{0.405} (0.006) & \textbf{0.390} (0.026) & \textbf{0.293} (0.006) & \textbf{0.398} (0.006) & \textbf{0.089} (0.026) & \textbf{0.312} (0.006) \\
 & DeepSeek & \textbf{0.411} (0.006) & \textbf{0.389} (0.016) & \textbf{0.284} (0.006) & \textbf{0.398} (0.006) & \textbf{0.121} (0.006) & \textbf{0.281} (0.006) \\
\bottomrule
\end{tabular}
\end{table}

\section{Full five-zone forecasting results}
\label{app:forecasting-full}

\subsection{Direct text fusion}

Table~\ref{tab:rawtext-full} reports all three quantile levels against
\textsc{NumOnly}, a separately trained control using the same architecture and
training procedure but with text features zeroed during training.
$\mathrm{PB}_{0.9}$ decreases in all five zones, whereas
$\mathrm{PB}_{0.1}$ increases in all five and $\mathrm{PB}_{0.5}$ in four of
five. This mirrors the semantic-alignment results: the forecasting value of
textual context is concentrated in upper-tail price risk rather than broad
distributional improvement.

Shuffle mean is the mean $\mathrm{PB}_{0.9}$ over $1{,}000$ test-time
permutations of textual context across the 131 held-out context-bearing test
days. Trained models are reused; only the day--context pairing is permuted,
while numerical inputs and targets remain fixed. Correct forecast-time
day--context pairing outperforms the Shuffle mean in all five zones, with a
median reduction of 20.7\%. Moreover, the Shuffle mean is worse than
\textsc{NumOnly} in all five zones, showing that mispaired textual context can
degrade forecasts rather than merely fail to improve them. Correct forecast-time
day--context pairing is significantly better than random re-pairing under the
aggregate permutation test, outperforming all $1{,}000$ permuted pairings
($p=1/1001$). Together with the semantic-alignment results, this supports the
forecasting value of correctly paired official operational text beyond price
history alone.

\begin{table}[htbp]
\centering
\caption{Direct text fusion against the \textsc{NumOnly} control and
shuffled-context control, five zones, 131 context-bearing test days each
(pinball loss, USD/MWh, at the released four-decimal precision).
$\downarrow$ columns are relative reductions (\%) of Direct from
\textsc{NumOnly} and from Shuffle mean at $\tau=0.9$.}
\label{tab:rawtext-full}
\scriptsize
\setlength{\tabcolsep}{2.6pt}
\begin{tabular}{lrrrrrrrrr}
\toprule
& \multicolumn{2}{c}{$\tau=0.1$}
& \multicolumn{2}{c}{$\tau=0.5$}
& \multicolumn{5}{c}{$\tau=0.9$} \\
\cmidrule(lr){2-3}
\cmidrule(lr){4-5}
\cmidrule(lr){6-10}
Zone
& \textsc{NumOnly} & Direct
& \textsc{NumOnly} & Direct
& \textsc{NumOnly} & Direct
& Shuffle mean
& $\downarrow$\textsc{NumOnly}
& $\downarrow$Shuffle \\
\midrule
COMED & 3.1988 & 3.4357 & 6.8241 & 7.1517 & 5.2348 & 4.8973 & 6.0941 & 6.4 & 19.6 \\
AEP   & 2.9475 & 3.2939 & 6.2434 & 6.6321 & 4.7835 & 4.4380 & 5.5876 & 7.2 & 20.6 \\
DOM   & 4.1392 & 4.5165 & 11.0238 & 10.7575 & 9.2139 & 8.4872 & 10.7006 & 7.9 & 20.7 \\
PSEG  & 3.3089 & 3.5489 & 7.0988 & 7.4453 & 5.4075 & 5.2073 & 6.8318 & 3.7 & 23.8 \\
BGE   & 3.5840 & 4.1484 & 9.4686 & 9.5586 & 7.6196 & 6.9271 & 9.5423 & 9.1 & 27.4 \\
\bottomrule
\end{tabular}
\end{table}

\subsection{Structured semantic fusion}

Table~\ref{tab:semantic-full} reports all 70 zone--model--LLM settings.
At $\tau=0.9$, structured semantic fusion improves every setting relative to
the separately trained \textsc{Cnum} control, with a median reduction of
$7.4\%$ (range $1.4$--$19.3\%$). The effect is weaker at the other quantiles:
64/70 settings improve at $\tau=0.5$, whereas only
24/70 improve at $\tau=0.1$. Forecasting value is
therefore strongest and most consistent for upper-tail price risk.

Table~\ref{tab:semantic-full} also reports the shuffled-context control.
For each trained model, textual context is randomly re-paired across the 131
held-out context-bearing test days $1{,}000$ times, while numerical inputs and
targets remain fixed. Correct forecast-time day--context pairing outperforms
the Shuffle mean in all 70/70 settings, with a median reduction of $13.9\%$.
The aggregate statistic outperforms all $1{,}000$ random re-pairings
($p=1/1001$), supporting that the gains depend on correctly matched
forecast-time textual context rather than text-derived features alone.

\begin{table}[htbp]
\centering
\caption{Structured semantic fusion for all 70 zone--model--LLM settings on
131 context-bearing test days (pinball loss, USD/MWh; lower is better).
\textsc{Frozen} is the frozen numerical baseline and \textsc{Cnum} the
separately trained control with the semantic vector zeroed during training.
Fusion uses the correct forecast-time day--context pairing; Shuffle is the mean
over $1{,}000$ test-time context re-pairings. The final two columns report
Claude Fusion minus \textsc{Cnum} at the other quantiles; negative favors
structured semantic information.}
\label{tab:semantic-full}

\scriptsize
\setlength{\tabcolsep}{1.8pt}

\begin{tabular*}{\linewidth}{
@{\extracolsep{\fill}}
llrrrrrrrr
@{}
}
\toprule
& &
\multicolumn{2}{c}{Numerical}
& \multicolumn{2}{c}{Claude Opus 5}
& \multicolumn{2}{c}{DeepSeek V4-Flash}
& \multicolumn{2}{c}{Claude, other $\tau$} \\
\cmidrule(lr){3-4}
\cmidrule(lr){5-6}
\cmidrule(lr){7-8}
\cmidrule(lr){9-10}

Zone
& Baseline
& \textsc{Frozen}
& \textsc{Cnum}
& Fusion
& Shuffle
& Fusion
& Shuffle
& $\Delta_{0.1}$
& $\Delta_{0.5}$ \\
\midrule

COMED
& TimesFM 2.5
& 4.1233 & 4.0277
& 3.6358 & 4.2001
& 3.6732 & 4.2970
& $+$0.1222 & $-$0.0354 \\

&
TimesFM 3.0
& 3.7222 & 3.6047
& 3.4027 & 3.8609
& 3.4160 & 3.8716
& $+$0.1203 & $-$0.0326 \\

&
Chronos-Bolt-Tiny
& 3.8887 & 3.7881
& 3.5255 & 3.9029
& 3.5229 & 3.9579
& $-$0.0052 & $-$0.0433 \\

&
Chronos-Bolt-Base
& 3.6671 & 3.5301
& 3.3962 & 3.7125
& 3.4100 & 3.7610
& $+$0.0560 & $-$0.0312 \\

&
Chronos-2
& 3.7438 & 3.6154
& 3.4640 & 3.8474
& 3.4765 & 3.8603
& $-$0.1128 & $+$0.0238 \\

&
Moirai-2
& 3.8633 & 3.7654
& 3.4646 & 4.0921
& 3.5058 & 4.1632
& $+$0.0494 & $-$0.0374 \\

&
Sundial
& 4.7661 & 3.8617
& 3.6059 & 4.0691
& 3.6257 & 3.9750
& $+$0.1192 & $-$0.0259 \\

\addlinespace[1pt]

AEP
& TimesFM 2.5
& 4.2411 & 4.0516
& 3.6287 & 4.1431
& 3.7786 & 4.1358
& $-$0.0564 & $-$0.0148 \\

&
TimesFM 3.0
& 3.8706 & 3.7702
& 3.4318 & 3.8026
& 3.5084 & 3.8051
& $-$0.0048 & $-$0.0060 \\

&
Chronos-Bolt-Tiny
& 4.0715 & 3.9453
& 3.5092 & 4.0159
& 3.5096 & 4.0646
& $-$0.0153 & $-$0.0545 \\

&
Chronos-Bolt-Base
& 3.7908 & 3.6280
& 3.3300 & 3.7776
& 3.3758 & 3.8253
& $+$0.0052 & $-$0.0343 \\

&
Chronos-2
& 3.7989 & 3.6524
& 3.2808 & 3.7383
& 3.3258 & 3.7722
& $-$0.0439 & $-$0.0126 \\

&
Moirai-2
& 3.9413 & 3.8216
& 3.5331 & 3.9468
& 3.5349 & 3.9660
& $-$0.0082 & $+$0.0012 \\

&
Sundial
& 4.6425 & 4.0567
& 3.8231 & 4.1385
& 3.9080 & 4.1207
& $+$0.0342 & $-$0.0024 \\

\addlinespace[1pt]

DOM
& TimesFM 2.5
& 8.1194 & 7.5317
& 7.2175 & 8.5693
& 7.4259 & 8.9055
& $+$0.0800 & $-$0.1190 \\

&
TimesFM 3.0
& 6.9558 & 6.9140
& 6.5728 & 7.6190
& 6.7584 & 7.9382
& $-$0.0864 & $-$0.1371 \\

&
Chronos-Bolt-Tiny
& 7.8417 & 7.4540
& 6.9945 & 8.0888
& 7.1968 & 8.4098
& $+$0.0556 & $-$0.1564 \\

&
Chronos-Bolt-Base
& 7.4483 & 7.1291
& 6.7142 & 7.4050
& 6.7671 & 7.6506
& $-$0.0380 & $-$0.0267 \\

&
Chronos-2
& 7.2247 & 7.0263
& 6.5210 & 7.2431
& 6.6039 & 7.4030
& $-$0.0275 & $-$0.1157 \\

&
Moirai-2
& 7.3162 & 6.9226
& 6.4897 & 7.5489
& 6.6172 & 7.9101
& $-$0.0104 & $-$0.1236 \\

&
Sundial
& 8.7192 & 7.5445
& 7.1805 & 8.1300
& 7.3345 & 8.3948
& $-$0.0180 & $-$0.0692 \\

\addlinespace[1pt]

PSEG
& TimesFM 2.5
& 4.7462 & 4.5028
& 3.6316 & 4.8476
& 3.8447 & 4.8606
& $+$0.0600 & $-$0.0637 \\

&
TimesFM 3.0
& 4.1577 & 4.0990
& 3.4458 & 4.4593
& 3.5172 & 4.2444
& $+$0.0954 & $-$0.0416 \\

&
Chronos-Bolt-Tiny
& 4.5423 & 4.3244
& 3.7156 & 4.8484
& 3.7908 & 4.7851
& $+$0.0365 & $-$0.0459 \\

&
Chronos-Bolt-Base
& 4.2154 & 3.9650
& 3.6289 & 4.2607
& 3.6753 & 4.2254
& $-$0.0340 & $-$0.0001 \\

&
Chronos-2
& 3.9860 & 3.6892
& 3.3003 & 4.2654
& 3.4271 & 4.3684
& $+$0.2009 & $-$0.0529 \\

&
Moirai-2
& 4.1734 & 4.0401
& 3.4322 & 4.3574
& 3.4915 & 4.2509
& $+$0.0242 & $-$0.0126 \\

&
Sundial
& 5.1508 & 4.4557
& 3.9284 & 4.8508
& 3.9715 & 4.8057
& $+$0.0845 & $-$0.0346 \\

\addlinespace[1pt]

BGE
& TimesFM 2.5
& 7.3484 & 7.0546
& 6.0522 & 7.3968
& 6.0698 & 7.6222
& $+$0.0285 & $-$0.1063 \\

&
TimesFM 3.0
& 6.1309 & 6.0823
& 5.5244 & 6.7056
& 5.6699 & 6.9312
& $-$0.0462 & $-$0.0553 \\

&
Chronos-Bolt-Tiny
& 6.9677 & 6.6735
& 5.8372 & 7.2193
& 5.9310 & 7.4830
& $+$0.0722 & $-$0.1170 \\

&
Chronos-Bolt-Base
& 6.3711 & 6.4523
& 5.8882 & 6.8994
& 5.8969 & 6.9826
& $-$0.0193 & $-$0.0548 \\

&
Chronos-2
& 5.9862 & 5.9018
& 5.4877 & 6.3647
& 5.5338 & 6.4618
& $+$0.0325 & $-$0.0782 \\

&
Moirai-2
& 6.4536 & 6.2549
& 5.5905 & 7.0253
& 5.7444 & 7.3657
& $+$0.0411 & $-$0.1066 \\

&
Sundial
& 7.6341 & 6.6445
& 5.9555 & 7.1357
& 6.0088 & 7.2778
& $+$0.0038 & $-$0.0667 \\

\bottomrule
\end{tabular*}
\end{table}

\end{document}